\documentclass[journal]{IEEEtran}

\ifCLASSINFOpdf
\else
   \usepackage[dvips]{graphicx}
\fi

\usepackage{graphicx}
\usepackage{url}
\usepackage{caption}
\usepackage{epstopdf}
\usepackage{newtxmath}
\usepackage{textcomp}
\usepackage{cite}
\usepackage{multicol}
\usepackage{etoolbox}
\usepackage{balance}
\usepackage{color}
\begin{document}

\title{Deep Learning for Content-based Personalized Viewport Prediction of 360-Degree VR Videos}

\author{Xinwei Chen, Ali Taleb Zadeh Kasgari, \IEEEmembership{Member, IEEE}, and Walid Saad, \IEEEmembership{Fellow, IEEE}

\vspace{-3em}
\thanks{Xinwei Chen is with Department of Control Science and Engineering, Zhejiang University, Hangzhou, China (e-mail: xwchen.zju@gmail.com).}
\thanks{Ali Taleb Zadeh Kasgari, is with Wireless@VT, Electrical and Computer Engineering Department Virginia Tech, Blacksburg, VA 24060 USA. (e-mail: alitk@vt.edu).}
\thanks{Walid Saad is with Wireless@VT, Electrical and Computer Engineering Department, Virginia Tech, Blacksburg, VA 24060 USA. (e-mail: walids@vt.edu).}
\thanks{This research was supported by the U.S. National Science Foundation under Grant IIS-1633363.}
}

\maketitle

\begin{abstract}
In this paper, the problem of head movement prediction for virtual reality videos is studied. In the considered model, a deep learning network is introduced to leverage position data as well as video frame content to predict future head movement. For optimizing data input into this neural network, data sample rate, reduced data, and long-period prediction length are also explored for this model. Simulation results show that the proposed approach yields 16.1\% improvement in terms of prediction accuracy compared to a baseline approach that relies only on the position data.
\end{abstract}

\IEEEpeerreviewmaketitle

\vspace{-1.5em}
\section{Introduction}

\IEEEPARstart{V}{irtual} reality (VR) is arguably one of the most exciting technologies for the coming decade \cite{vrintro2}. Major virtual reality headset producers such as HTC VIVE, Oculus Go, and PlayStation VR are developing advanced head mount display (HMD) with high resolution and high refresh rate in order to enable a plethora of innovative VR applications. Meanwhile, 360\textdegree~video providers like YouTube and Facebook can provide ultra-high-resolution (up to 16K) with $120$~Hz frame rate panoramic videos to users for a better immersion experience. However, to deliver VR and 360\textdegree~video content wireless over the Internet, VR providers must transmit their content in a bandwidth-efficient manner over capacity-constrained wireless networks \cite{6G}. In particular, the whole VR video is downloaded and delivered to HMD when watching, but the observable area in the device, called field of view (FoV) or viewport, is only part of the whole video downloaded. While a VR user can adjust its orientation by changing its pitch, yaw, and roll, the viewport displayed in HMD is based on the head movement and FoV. Therefore, most of the VR video delivered to HMD is not displayed in the viewport. 

The FoV defines the extent of the area observable by a VR user and this area is always a fixed parameter of a VR headset (typically 110\textdegree). To ensure a good viewing experience, the video delivered is often transmitted in extremely high resolution, typically 4K $(3840 \times 2160)$, while in HMDs the VR users only see a small part of the video. Thus, in order to save network bandwidth, it is necessary to develop new approaches that allow a VR content provider to deliver only the viewport to the HMD of its users.

Tile-based panoramic video streaming \cite{tile} is a special type of panoramic video streaming which can deliver part of the video content, instead of the whole video. In order to deliver only the viewport content with tile streaming, it is imperative to develop new approaches to predict head movement for a VR user. In this way, VR content providers can only deliver the video portions which the user is most likely to view in high quality while the remaining portions are ignored or delivered in low quality. 

Despite tremendous efforts devoted to the prediction of user's viewport in 360\textdegree~videos \cite{qian, duanmu,  aladagli, fan, assens, nguyen, xu}, there are no works considering both head movement and content of the video for predicting future viewport of the video. By taking both head movement and video content into account, the VR system can be designed to outperform previous works in predicting the viewport of a 360\textdegree~video for the user. 

\vspace{-1.2em}
\subsection{Related Works}

Some recent works such as in \cite{qian, duanmu} have studied predicting head movement using only history head orientation data. They propose methods such as simple average \cite{qian}, linear regression \cite{duanmu}, and weighted linear regression \cite{qian} to predict future head movement. However, these approaches in \cite{ qian, duanmu } use naive models and ignore video content's relation to future movement, thus are less accurate. Other existing works such as in \cite{aladagli, fan,assens, nguyen, xu} combine both the video content features and orientation of HMD to predict the future head movement. In \cite{aladagli}, the authors use a pre-trained saliency model to predict head movement. The work in \cite{fan} integrated saliency, motion map, and prior head orientation to further improve the prediction accuracy. The authors in \cite{assens} introduce {\em SaltiNet} for scan-path prediction trained on 360\textdegree~images. Moreover, the work in \cite{nguyen} proposes panoramic saliency based on the user's fixation along with the user's head movement history for prediction. The authors in \cite{xu} combine saliency map, gaze displacement and history scan-path for a better prediction performance.

Although the works in \cite{aladagli, fan, assens, nguyen, xu} use both video saliency and history head orientation to predict future head movement, these prior works do not directly treat video frame content in much detail. Since VR videos contain various scenes, and further, each scene has different regions of interest for users. In fact, remarkably, to date, none of the prior works has investigated whether video frame content can contribute to the performance of head movement prediction. This study will generate fresh insight into this frame content and head trajectory-based architecture.

\vspace{-1.2em}
\subsection{Contributions}
The main contribution of this paper is a novel framework for predicting the future head movement of VR users while watching a 360\textdegree~video. To the best of our knowledge, this is the first work that considers both position-related data and content-related data VR user's head movement prediction. First, we propose a deep learning network to predict future head movement, this network employs a convolutional neural network (CNN) as feature extractor of content-related data and long short-term memory (LSTM) as the encoder of position-related data to predict the future movement of a user. The position-related data includes the VR user's head orientation, while the content-related data is each VR video frame itself. Simulation results using real data show that the proposed CNN-LSTM architecture can achieve high prediction accuracy, in particular, the proposed content-based approach can yield up to 16.1\% compared to a baseline that relies only on position.

\section{Problem Statement and Proposed Approach}
In this section, we will describe our approach toward head movement prediction.\

\vspace{-1em}
\subsection{Problem Definition}
The head movement prediction problem is formulated as follows: We use a set $\mathcal{V}_{1:t} = \{v_1, v_2,...,v_t\}$ to represent a sequence of 360\textdegree~ video frames where $v_i\in\mathcal{V}$ corresponds to frame $i$. In VR, the standard format of a VR video frame is in equirectangular projection{\cite{equi}}. We use a set $\mathcal{L}_{1:t} = \{\boldsymbol{l}_1, \boldsymbol{l}_2,..., \boldsymbol{l}_t\}$ to represent the head orientation in $ \mathcal{V}_{1:t} $ frames where $\boldsymbol{l}_i\in\mathcal{L}$ is coordinates of head rotation. Here $\boldsymbol{l}_i=\left[ q_0, q_1, q_2, q_3 \right]$, to be specific, is named \emph{quaternion} which has a hyper-complex number of rank 4, and often used in VR systems to describe rotations\cite{quat}. Based on the above formulation, movement prediction aims to regress the future movement coordinates corresponding to the future $T$ frames. The predicted movement set is defined as $\mathcal{L}_{t+1:t+T} = \{\boldsymbol{l}_{t+1}, \boldsymbol{l}_{t+2},..., \boldsymbol{l}_{t+T}\}$.

The future head movement of a VR user is related to multiple factors. On the one hand, users are more likely to be attracted by some features in the frame. On the other hand, history movement is also a key factor to predict the future movement of a user because of different movement patterns. 

\begin{figure}[!t]
    \centering
    \includegraphics[width=\columnwidth]{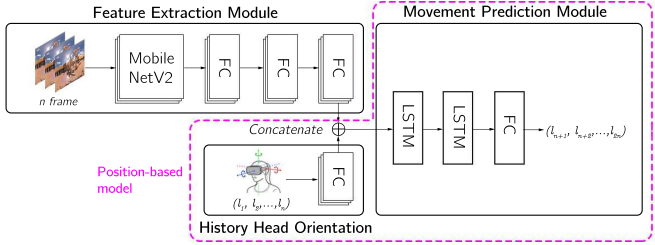}
    \caption{The architecture of our proposed deep neural network.}
    \label{fig:model_structure}
    \vspace{-1.3em}
\end{figure}

Based on the above observations, we can pose the head movement prediction problem as a non-linear regression problem, in which the history head movement and corresponding frames are independent variables and future head movement are dependent variables. We use a deep neural network for this prediction problem (as shown in Fig.~\ref{fig:model_structure}). Our network includes a feature extraction module and a movement prediction module. This model has the following advantages: i) It can exploit important information from the video which is neglected in the previous model, ii) It can be implemented over a mobile device because of its lightweight in its architectures, and iii) It can effectively learn both short and long term movement of users because of the merit of LSTM. We will elaborate on the model in the next section.

\vspace{-1em}
\subsection{Movement Prediction Network}
\subsubsection{Feature extraction module} 
Different methods have been used to extract features from the image. One suitable approach is to use neural networks, particularly, CNNs to extract image features. CNNs are suitable for this problem because they showed a tremendous success in end-to-end learning tasks such as object detection and image classification. We adopt a deep multi-level network, whose architecture is based on MobileNetV2 \cite{mobilenet}. MobileNetV2 is a very effective feature extractor based on an inverted residual structure that can be implemented on the user's mobile devices easily due to its lightweight. A pre-trained MobileNetV2 model follows with three fully connect layers to extract the feature of the image. This network extracts 32 features for each frame. The output of CNN is now defined as follows:
\vspace{-0.5em}
\begin{equation}
\vspace{-0.5em}
f_i = h_1(v_i),
\end{equation}

where function $h_1(\cdot)$ represents the input-output function of MobileNetV2.

\subsubsection{Movement prediction module}
The function of this network is to predict where the VR user is more likely to look at in the future video frames given a sequence of concatenated observed video frames and position-related data. The movement prediction network, which is based on a recurrent neural network (RNN) \cite{MLpaper}, is suitable to learn useful information from a time series of video frames. In view of the good performance of LSTM networks in motion tracking and its ability to learn long-term dependencies, we choose an LSTM-based RNN for predicting future head movements.

The network takes as inputs the features of $n$ past video frames as well as the head orientation data in a sliding window. Then, we can define the output set of stacked LSTM $\mathcal{L}_{t+1:t+T}$ as:
\vspace{-0.5em}
\begin{equation}
\vspace{-0.5em}
\boldsymbol{l}_{t+1}, \boldsymbol{l}_{t+2},..., \boldsymbol{l}_{t+T} = h_2(f_1+\boldsymbol{l}_{1}, f_2+\boldsymbol{l}_{2},..., f_n+\boldsymbol{l}_{n}),
\end{equation}
where function $h_2(\cdot)$ represents the input-output function of stacked LSTM. In addition, the function $h_2$ is stacked LSTMs with 2 LSTM layers and one fully connect layer, each with 256 neurons.

Hence, for this model, our objective is to predict the future $T$ movement using deep learning methods. The model proposed above can minimize the prediction error while meeting the memory constraint on mobile device and delay requirement of the VR system. 

\vspace{-1em}
\subsection{Performance Metric}
We propose to use the viewing angle between the predicted orientation and the actual one to measure the performance of head movement prediction. A small angle implies a more accurate orientation prediction, and, hence, we use the mean angle error (MAE) to evaluate the performance. For head orientation in frame $i$, we convert the quaternion expression to a Euler angle $(x_i, y_i)$ which indicate the latitude and longitude of head orientation on a 3D-Sphere where $x_i\in[-180, 180], y_i\in[-90, 90]$. We define that $(x_i, y_i)$ as actual head orientation and $(\hat{x_i}, \hat{y_i})$ as predicted one. Here the angle error $\sigma$ at $i$ frame can be defined as: 
\vspace{-0.5em}
\begin{equation}
\vspace{-0.5em}
\sigma_{i} = \arccos \left( \sin x_i\sin \hat{x_i} + \cos x_i\cos \hat{x_i} \cos \lvert y_i-\hat{y_i} \rvert \right),
\end{equation}

We also use cumulative distribution function (CDF) of all head orientation for performance evaluation. A higher CDF curve corresponds to a method with smaller MAE. 

In summary, we have introduced a feature extraction module for video content and a movement prediction module for combining content-related features and position-related feature to predict future movement of VR users. Then, we use MAE and CDF to evaluate our proposed model. By adding the content-related features to our system, we can implicitly predict future points of interest in the video. This prediction gives an advantage to our proposed method over previous studies for head movement prediction \cite{qian, duanmu}.

\vspace{-1em}
\subsection{Exploiting Predictions in a Wireless Network}
 As discussed in \cite{cellularconnected}, deploying a VR system over a wireless network will potentially strain the capacity and bandwidth of that network, thus requiring new techniques to improve resource management.  One promising approach to overcome this challenge, is by enabling the VR system to predict the head movement of VR users, as proposed here. In particular, by performing such head movement prediction, instead of using a very large bandwidth to transmit the entirety of a VR content to a user (e.g., the entirety of a 360\textdegree~video),  the wireless network can now adaptively decide on what portion of the VR content to deliver to its users at any given time thus saving significant bandwidth and resources. Also, performing predictions of a VR user's head movement can enable the wireless network to pre-fetch and cache the predicted parts of the content at the edge of the network, thus reducing latency and bandwidth usage, particularly at backhaul links. Clearly, the proposed prediction scheme can be leveraged in all of these scenarios.

\vspace{-1em}
\section{Simulation Results and Discussions}

\subsection{Dataset}
There exist a few datasets of head orientation for VR users. We explore one public head movement dataset for 360\textdegree~videos \cite{data1}. This dataset includes 5 videos viewed by 59 users. Each head orientation is timestamped and corresponding to a video frame. The head orientation is recorded as a quaternion, a four-tuple mathematical representation of head orientation with respect to a fixed reference point. We use the quaternion expression as the input and output of our model.

\vspace{-1em}
\subsection{Simulation Setup}
\begin{figure}[!t]            
    \centerline{\includegraphics[width=\columnwidth]{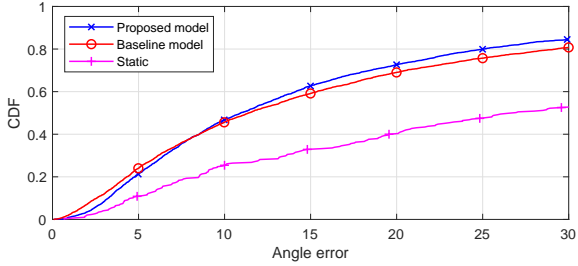}}
    \caption{Comparison of the CDF of the angle error.}
    \label{fig:lstmvsimg}
    \vspace{-1.5em}
\end{figure}

\begin{figure}[!t]
    \centerline{\includegraphics[width=\columnwidth]{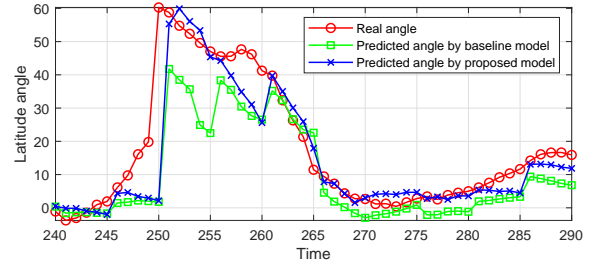}}
    \caption{Head latitude angle prediction for the proposed model and the baseline model}
    \label{fig:angle_compare}
    \vspace{-1.5em}
\end{figure}

\begin{figure}[!t]
    \centerline{\includegraphics[width=\columnwidth]{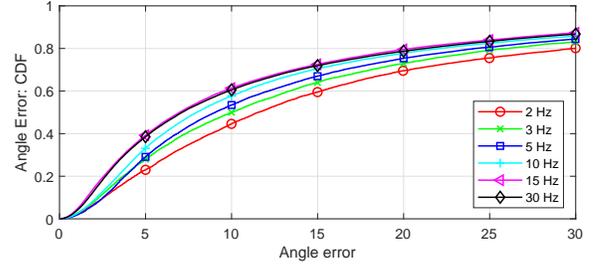}}
    \caption{Effect of input sample rate on the angle error.}
    \label{fig:samplerate}
    \vspace{-1.5em}
\end{figure}

We downsample one frame from every five frames for model training and evaluation. Hence, the interval between two frames in our experiments is $\frac{1}{6}$ seconds. This setting can reduce the memory consumption needed for training and evaluating our model,  yet it makes the head movement prediction more challenging compared to the original frame-rate. In what follows, the frames mentioned correspond to the downsampled ones. We use five history head orientation data to predict head movement in the next five frames. In other words, we use the first 1 second to predict the next second.

The model is implemented using the TensorFlow framework. We train our network with the following hyper-parameters setting: mini-batch size(32), Adam optimizer, learning rate(0.001), and decay(0.0005). Our training model is a generalizable model that applies for all users, as per the process that we explain next. In our simulations, for each video, we randomly select 70\% of the users' movement behavior for training data, out of which $10\%$ are used for validation and $20\%$ for testing. The model is trained for 500 epochs and is checkpointed after the minimum validation loss.

\vspace{-1em}
\subsection{Results}
In Fig.~\ref{fig:lstmvsimg}, we compare the model we proposed with a baseline, position-based model. The position-related model (as shown in Fig.~\ref{fig:model_structure}) only takes $5$ past frame head orientation data points in one second denoted by $\mathcal{L}_{1:5}$ into a stacked LSTM to predict future movement as captured by $\mathcal{L}_{t+1:t+5}$. Fig.~\ref{fig:lstmvsimg} shows that the proposed model yields up to 16.1\% improvement in MAE compared to the baseline. Fig.~2 also shows that both the proposed model and position-based model perform far better than static prediction (the system assumes the user does not move), which validates the necessity of our prediction. Fig.~\ref{fig:angle_compare} shows a comparison between the real latitude angle and predicted and the predicted angles. From this figure, we observe that our proposed model achieves higher accuracy than the baseline. This improvement in error stems from the fact that MobileNetV2 can extract the user's regions of interest from the video content, and, hence, it enables the LSTM to use these video features and combines them with history trajectory data to enhance the prediction accuracy.

Fig.~\ref{fig:samplerate} shows how prediction accuracy changes as the sample rate vary. From Fig.~\ref{fig:samplerate}, we can see that, as the sample rate increases, the prediction accuracy increases significantly around $2$~Hz to $15$~Hz, and at $15$~Hz to $30$~Hz the increase becomes slower. This is because, as the sample rate increases, more data will be input to our network and, thus, more features will be extracted. Fig.~\ref{fig:samplerate} also shows that when the sample rate reaches $15$~Hz, the prediction accuracy remains constant or decreases as the sample rate increases. This is because by increasing sampling frequency beyond a certain value, input data to the model becomes mostly redundant. This redundancy will increase the model complexity as well as the time to execute our algorithm, but it cannot improve model accuracy. Thus, such sample rate cannot meet the low latency requirement for HMD operation and guarantee user's quality of experience. Under a computation memory constraint and time, we need to reduce the history data transmitted to the model and also achieve good prediction accuracy. Taking the sample rate as $5$~Hz can be a good compromise of these two factors.

\begin{figure}[!t]
    \centerline{\includegraphics[width=\columnwidth]{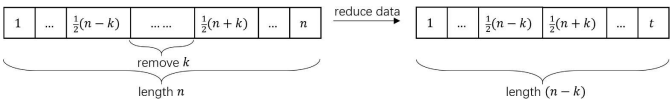}}
    \caption{Reduce $k$ data points in a time window of length $n$}
    \label{fig:reducedata}
    \vspace{-1.7em}
\end{figure}

\begin{figure}[!t]
    \centerline{\includegraphics[width=\columnwidth]{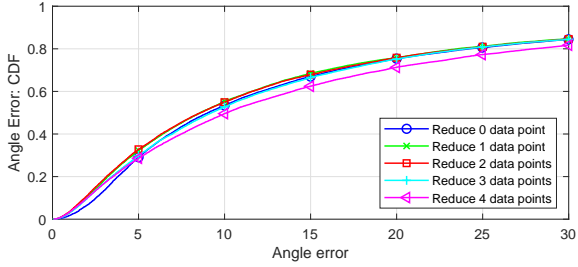}}
    \caption{Effect of reduced input dataset on the angle error.}
    \label{fig:cutlength}
    \vspace{-1.7em}
\end{figure}

Fig.~\ref{fig:cutlength} shows the performance of our proposed model on a reduced dataset  with a sample rate of $5$~Hz. We remove $k$ data points in the middle of a time window of length n, as shown in Fig.~\ref{fig:reducedata}. From Fig.~\ref{fig:cutlength}, we can see that the prediction accuracy varies little as the reduced data length increases from 0 to 3. This implies that we can save computation memory and upload bandwidth by reducing the input data and achieve the same performance as before. 

\begin{figure}[!t]
    \centerline{\includegraphics[width=\columnwidth]{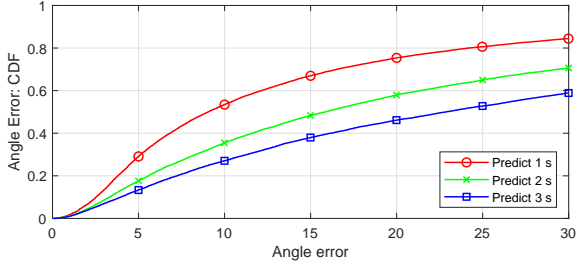}}
    \caption{Effect of input duration on the angle error.}
    \label{fig:predictlength}
    \vspace{-1.5em}
\end{figure}

Fig.~\ref{fig:predictlength} shows how our model's performance changes as the prediction length vary. From Fig.~\ref{fig:predictlength}, we observe that it is difficult to predict long-term movement. This is due to the fact that future movements over an in long period are less relevant to the current movement, and, hence the prediction length should be limited for better performance.

\vspace{-0.5em}
\section{Conclusion and Future Work}

In this paper, we have proposed neural network architecture for predicting the head movement of VR users. First, we have introduced a network that uses CNN to extract the features of video content, and an LSTM to predict the future movement of users. Simulation results have shown that the proposed approach yields an improvement in prediction accuracy compared to the position-related only one. Next, we have found that downsample or reduced input dataset can meet the constraint of computation memory and upload bandwidth while maintaining the prediction accuracy. Extensive experiments also validate the effectiveness of our approaches. In future work, by considering motions between video frames, training with a larger dataset and using eye-tracking HMD may further boost the performance. Another important future work is to conduct actual experiments with real VR users to further validate our approach.
\vspace{-1em}
\bibliographystyle{IEEEtran}
\bibliography{paper.bbl}

\end{document}